\documentclass[runningheads]{llncs}
\usepackage[T1]{fontenc}
\usepackage{graphicx}
\usepackage{amsmath,amssymb,amsfonts}

\usepackage{textcomp}

\usepackage{float}
\usepackage{rotating}
\usepackage{longtable}
\usepackage{adjustbox}
\usepackage{booktabs}
\usepackage{multicol}
\usepackage{multirow}
\usepackage{mdframed}  

\usepackage{listings}

\usepackage[ruled,vlined,linesnumbered]{algorithm2e}

\usepackage[hidelinks]{hyperref}

\usepackage{cite}
\usepackage{subcaption}
\usepackage{lipsum}

\newmdenv[
  backgroundcolor=white,
  linecolor=black,
  roundcorner=5pt,
  innerleftmargin=5pt,
  innerrightmargin=5pt,
  innertopmargin=5pt,
  innerbottommargin=5pt,
  linewidth=1.5pt,
  skipabove=4pt,
  skipbelow=4pt,
  splittopskip=4pt,
  splitbottomskip=4pt
]{mySummaryBox}

\newboolean{showcomments}
\setboolean{showcomments}{true} 

\ifthenelse{\boolean{showcomments}}{%
  \newcommand{\nb}[2]{%
    \fcolorbox{black}{yellow}{\bfseries\sffamily\scriptsize #1}~%
    {\sf\small$\blacktriangleright$ \textit{#2} $\blacktriangleleft$}%
  }%
}{%
  \newcommand{\nb}[2]{}%
}

\begin{document}
\title{FAROS: Robust Federated Learning with Adaptive Scaling against Backdoor Attacks}
%
%
\author{Chenyu Hu\inst{1}\thanks{This study is conducted during the visit to Waseda University} \and
Qiming Hu\inst{2} \and
Sinan Chen\inst{3} \and
Nianyu Li\inst{4} \and
Mingyue Zhang\inst{1} \and
Jialong Li\inst{5}\thanks{Corresponding author. E-mail: lijialong@fuji.waseda.jp}}

\authorrunning{C. Hu et al.}

\institute{Southwest University, Chongqing, China, 400715
\and University of Electronic Science and Technology of China, Chengdu, China, 611731
\and Kobe University, Kobe, Japan, 657-8501
\and Zhongguancun Laboratory, Beijing, China, 100085
\and Waseda University, Tokyo, Japan, 169-8555}

\maketitle              
\begin{abstract}
Federated Learning (FL) enables multiple clients to collaboratively train a shared model without exposing local data. 
However, backdoor attacks pose a significant threat to FL. These attacks aim to implant a stealthy trigger into the global model, causing it to mislead on inputs that possess a specific trigger while functioning normally on benign data. 
Although pre-aggregation detection is a main defense direction, existing state-of-the-art defenses often rely on fixed defense parameters.
This reliance makes them vulnerable to single-point-of-failure risks, rendering them less effective against sophisticated attackers.
To address these limitations, we propose FAROS, an enhanced FL framework that incorporates Adaptive Differential Scaling (ADS) and Robust Core-set Computing (RCC). 
The ADS mechanism adjusts the defense's sensitivity dynamically, based on the dispersion of uploaded gradients by clients in each round. This allows it to counter attackers who strategically shift between stealthiness and effectiveness.
Furthermore, the RCC effectively mitigates the risk of single-point failure by computing the centroid of a core set comprising clients with the highest confidence.
We conducted extensive experiments across various datasets, models, and attack scenarios. The results demonstrate that our method outperforms current defenses in both attack success rate and main task accuracy. 

\end{abstract}
\section{Introduction}
\label{sec:introduction}
Federated Learning (FL) \cite{DBLP:conf/aistats/McMahanMRHA17,DBLP:journals/corr/KonecnyMYRSB16} is a distributed machine learning paradigm aimed at overcoming the challenge of the data island problem while preserving data privacy. 
This framework consists of a central server that manages a set of clients (e.g., mobile phones, IoT devices). The primary objective is to collaboratively train a global model by leveraging data from all clients without requiring them to share their local datasets. 
Specifically, in the first round, the server distributes the initial global model to all clients. Each client then performs local training on its private data to generate model updates (commonly in the form of gradients). Subsequently, these gradients uploaded back to the server, which aggregates them using an aggregation rule, such as Federated Averaging (FedAvg) \cite{DBLP:conf/aistats/McMahanMRHA17}, to refine the global model. 
This process continues in cycles until the performance of the global model converges \cite{DBLP:journals/tist/YangLCT19}. 

Despite the significant advantages of FL, its distributed nature makes it highly susceptible to various attacks \cite{DBLP:journals/cn/ZhangWLY25,DBLP:journals/tifs/BaiZZYH25,DBLP:conf/ccnc/SanonRLS24,DBLP:conf/aistats/FraboniVL21,DBLP:journals/tifs/MaHWQWM25,DBLP:journals/tbd/XiaoTLJL24,DBLP:journals/tifs/YangMLLLKLD25,DBLP:conf/cvpr/XieFG25,DBLP:conf/icml/BhagojiCMC19,DBLP:conf/uss/FangCJG20,DBLP:conf/nips/ZhangJCLW23,DBLP:conf/iclr/XieHCL20}, particularly Model Poisoning Attacks (MPA). In MPA, malicious clients can submit carefully crafted malicious gradients to the server, intending to compromise the availability of the global model. 
Among the various types of MPA, backdoor attacks \cite{DBLP:journals/tbd/XiaoTLJL24,DBLP:conf/icml/BhagojiCMC19,DBLP:conf/uss/FangCJG20,DBLP:conf/nips/ZhangJCLW23,DBLP:conf/iclr/XieHCL20,DBLP:conf/aistats/BagdasaryanVHES20,DBLP:conf/icml/ZhangPSYMMR022,DBLP:conf/sp/LiYHLWFS23} pose a more severe threat to the global model due to their stealthiness. The objective of the attacker is not to comprehensively degrade the global model, but rather to implant a backdoor. This manipulation ensures the global model produces a specific output only for inputs containing a particular trigger. In contrast, its performance on all other benign inputs remains indistinguishable from that of a benign model. This attack paradigm is difficult for defense mechanisms to detect. 

To counter backdoor attacks in FL, various defenses  \cite{DBLP:journals/access/ChenFW25,ACM:journals/jisa/Hu25,DBLP:journals/access/MasudaKKTH23,DBLP:journals/access/GhaziFN25,DBLP:journals/access/YeoL25} have been proposed. Based on their stage of action in FL, these defenses can be categorized into three types: pre-aggregation detection \cite{DBLP:journals/access/YeoL25,DBLP:journals/access/MasudaKKTH23,DBLP:journals/access/GhaziFN25,DBLP:journals/access/Almuseelem25,DBLP:conf/iccv/HuangLCS023,DBLP:journals/tifs/HuangLYGCSCN25,DBLP:conf/uss/NguyenRCYMFMMMZ22}, robust aggregation \cite{DBLP:journals/tsp/PillutlaKH22,DBLP:conf/icml/YinCRB18,DBLP:conf/aaai/OzdayiKG21}, and post-aggregation mitigation \cite{DBLP:journals/corr/abs-1911-07963,DBLP:journals/corr/abs-2011-01767}. 
Pre-aggregation detection aims to directly identify and filter out malicious gradients by analyzing uploaded gradients before aggregation, has become the main direction. However, early defenses were often constrained by strict assumptions, which limited their application in real-world scenarios.
For instance, distance-based aggregation rules like Multi-krum \cite{DBLP:conf/nips/BlanchardMGS17} require that the proportion of malicious clients does not exceed a threshold. 
FLTrust \cite{DBLP:conf/ndss/CaoF0G21} requires the server to maintain a dataset to bootstrap trust, a requirement that is often difficult to satisfy. 
Consequently, recent defenses \cite{DBLP:conf/iccv/HuangLCS023,DBLP:journals/tifs/HuangLYGCSCN25,DBLP:conf/uss/NguyenRCYMFMMMZ22} have shifted towards analyzing the features of gradients more broadly. 
For instance, Scope \cite{DBLP:journals/tifs/HuangLYGCSCN25} utilizes normalization, differential scaling, and clustering to amplify and expose hidden backdoor features within malicious gradients, providing an effective defense that does not rely on strict assumptions. 

Although Scope has shown significant success in defending against backdoor attacks, we identify two potential limitations when confronting more sophisticated attackers \cite{DBLP:journals/iotj/WangCZHWDLL25,DBLP:conf/uss/PanZWXJY20}.
First, Scope's differential scaling relies on a fixed scaling factor. 
A sophisticated attacker may dynamically adjust their attack strategy to balance between effectiveness and stealthiness. 
A fixed scaling factor struggles to adapt to such dynamic strategies.
If the factor is too high, it may misclassify benign gradients in heterogeneous data settings as malicious. 
On the other hand, if the factor is too low, it may fail to effectively amplify the features of a stealthy backdoor, thereby weakening the defense.
Second, the design of its algorithm makes it highly sensitive to the selection of its initial point. 
The algorithm uses a single gradient as the starting point for cluster expansion, which introduces a single point of failure. \cite{DBLP:journals/access/AliHJ25,DBLP:journals/access/NairNRN25}. 
In heterogeneous data environments, this dominant gradient can be erroneously selected, misleading results. 

In this work, we propose a new FL method called FAROS.
We introduce the Adaptive Differential Scaling (ADS). 
ADS dynamically sets the scaling factor by evaluating the degree of dispersion among all client gradients in the current round. 
When the gradient distribution is concentrated (suggesting a stealthy backdoor attack), ADS employs a higher scaling factor to amplify subtle backdoor features. 
Conversely, when the distribution is dispersed (indicative of a more aggressive backdoor attack or data heterogeneity), it adopts a more conservative factor to avoid erroneously penalizing benign yet anomalous clients. 
Additionally, to resolve the single-point-of-failure risk, we design the Robust Core-set Computing (RCC). 
Unlike reliance on a single dominant gradient, RCC first identifies a group of clients with the highest mutual similarity. 
This group is considered the most likely to be benign and is designated as the stable core-set.
Then calculates the centroid of this core-set and uses it as the reliable starting point for clustering, significantly enhancing stability under high heterogeneity and sophisticated attackers. 

Our main contributions are summarized below.
\begin{itemize}
    \item We propose FAROS, a novel backdoor defense framework for FL. By introducing dynamic and consensus mechanisms, our framework effectively addresses the core limitations of static parameters and single-point failures inherent in current state-of-the-art defenses. 

    \item We design an ADS mechanism that adaptively adjusts the defense's sensitivity based on the degree of dispersion among client gradients in each round. This enables FAROS to effectively counter sophisticated attackers who dynamically switch between stealthiness and effectiveness.

    \item We conduct extensive experiments to evaluate the performance of FAROS. Comprehensive comparisons across various datasets, models, and backdoor attacks, along with tests under different FL settings and ablation studies, demonstrate that our FAROS outperforms existing defenses.
\end{itemize}
The rest of this paper is structured as follows: 
Section \ref{sec: RelatedWork} provides the background of backdoor attacks and defenses in FL. 
Section \ref{sec: system} demonstrates the system setting, threat model, and defense goals.  
Section \ref{sec: method} introduces our method. 
Section \ref{sec: result} evaluates our method. 
Section \ref{sec: conclusion} concludes our work and provides our future work.

\section{Related work}
\label{sec: RelatedWork}
For Federated Learning (FL) \cite{DBLP:journals/corr/KonecnyMYRSB16}, Model Poisoning Attacks (MPA) \cite{DBLP:conf/icml/BhagojiCMC19,DBLP:conf/uss/FangCJG20,DBLP:conf/nips/ZhangJCLW23,DBLP:conf/iclr/XieHCL20,DBLP:conf/aistats/BagdasaryanVHES20,DBLP:conf/icml/ZhangPSYMMR022,DBLP:conf/sp/LiYHLWFS23,DBLP:journals/access/CuiDJZHY23,DBLP:journals/access/AlmutairiB25} pose a significant security threat. In such attacks, an attacker, i.e., a malicious client, manipulates the model gradients sent to the server to achieve a malicious objective. Based on the attacker's objective, MPA can be broadly categorized into two types: untargeted attacks and targeted attacks. Untargeted attacks \cite{DBLP:conf/icml/BhagojiCMC19,DBLP:conf/uss/FangCJG20} aim to comprehensively corrupt the global model's usability, for instance, by impeding convergence or degrading its overall performance. 
Among targeted attacks \cite{DBLP:conf/nips/ZhangJCLW23,DBLP:conf/iclr/XieHCL20,DBLP:conf/aistats/BagdasaryanVHES20,DBLP:conf/icml/ZhangPSYMMR022,DBLP:conf/sp/LiYHLWFS23,DBLP:conf/nips/WangSRVASLP20}, backdoor attacks are the most representative attacks. The objective of a backdoor attack is not to directly degrade performance but rather to inject a backdoor into the global model. This causes the global model to produce a specified incorrect output when presented with inputs containing a specific trigger while behaving normally on clean inputs, thus achieving a high degree of stealth. This research also focuses on backdoor attacks. 

\subsection{Backdoor Attacks in FL}
Based on the attack's capabilities, backdoor attacks can be classified into black-box attacks \cite{DBLP:conf/iclr/XieHCL20,DBLP:conf/nips/WangSRVASLP20,DBLP:conf/nips/ZhangJCLW23,DBLP:conf/sp/LiYHLWFS23} and white-box attacks \cite{DBLP:conf/aistats/BagdasaryanVHES20,DBLP:conf/icml/ZhangPSYMMR022,DBLP:conf/nips/WangSRVASLP20}. In black-box attacks \cite{DBLP:conf/iclr/XieHCL20,DBLP:conf/nips/WangSRVASLP20,DBLP:conf/nips/ZhangJCLW23,DBLP:conf/sp/LiYHLWFS23}, the attack's capabilities are strictly limited; they are confined to manipulating the local training dataset, and they cannot directly modify or manipulate model parameters after training. The term black-box is used because, from the attacker's perspective, the server-side aggregation algorithm, defense mechanisms, and the state of other clients are entirely unknown. This threat model is considered more realistic as it does not require the attacker to possess knowledge. Due to this lower barrier to attack, black-box backdoor attacks are considered a realistic attack in FL scenarios. 
A typical black-box attack is DBA \cite{DBLP:conf/iclr/XieHCL20}, which achieves stealth by decomposing a single global trigger into distinct local triggers. Each malicious client then trains using only its unique trigger, making it difficult for the server to detect the attack by analyzing any single client's gradients. Another approach leverages data distribution, such as the Edge-case Attack \cite{DBLP:conf/nips/WangSRVASLP20}, which uses edge-case data that is rare in benign clients' datasets to carry the backdoor task. The primary motivation is that, due to the lack of such data among benign clients, the backdoor is less likely to be forgotten by normal updates during the aggregation process. More advanced attacks introduce dynamic and adaptive mechanisms. For example, A3FL \cite{DBLP:conf/nips/ZhangJCLW23} abandons a fixed trigger, instead continually adjusting the trigger pattern via an optimization method to better resist defenses, thus improving the durability of the backdoor. Similarly dynamic, 3DFed \cite{DBLP:conf/sp/LiYHLWFS23} implants an indicator into the backdoor model and determines if the backdoor has been successfully injected by checking the indicator's status in the global model returned from the server. Based on this feedback, the attacker can dynamically adjust the magnitude of the backdoor to achieve a balance between attack effectiveness and stealth. 

In white-box attacks \cite{DBLP:conf/aistats/BagdasaryanVHES20,DBLP:conf/icml/ZhangPSYMMR022,DBLP:conf/nips/WangSRVASLP20}, the attacker possesses more complete control. They can not only manipulate local data, but more critically, are able to directly manipulate model parameters after local training. This powerful capability allows attack strategies to develop in two different directions: one is to maximize the attack's impact, and the other is to enhance stealth through sophisticated constraints. 
Among maximization strategies, Model Replacement \cite{DBLP:conf/aistats/BagdasaryanVHES20} is a direct approach that attempts to dominate the aggregation stage by significantly amplifying the magnitude of the backdoor model, thus covering the global model with the backdoor model. Neurotoxin \cite{DBLP:conf/icml/ZhangPSYMMR022} seeks the durability of the backdoor by projecting the backdoor gradient onto an area not used by benign clients, thereby preventing the backdoor from being eliminated by subsequent benign updates. 
In contrast, other white-box attacks focus more on stealth to evade detection. For example, Constrain-and-scale \cite{DBLP:conf/aistats/BagdasaryanVHES20} introduces a regularization term into the local loss function, which penalizes the deviation between the backdoor model and the previous global model, thus making it closer to benign models. The Edge-case PGD attack \cite{DBLP:conf/nips/WangSRVASLP20} is a more complex attack that leverages edge-case data to ensure the effectiveness of the backdoor task, while on the other hand, it employs the Projected Gradient Descent (PGD) \cite{DBLP:conf/iclr/MadryMSTV18} to strictly constrain the backdoor model within a neighborhood of the previous global model, thereby minimizing deviation and ensuring a high degree of stealth. 

\subsection{Backdoor Defenses in FL}
To counter the threat of backdoor attacks in FL, various defenses have been propsed \cite{ACM:journals/jisa/Hu25,DBLP:journals/access/MasudaKKTH23,DBLP:journals/access/GhaziFN25,DBLP:journals/access/YeoL25,DBLP:journals/access/Almuseelem25}. These defenses can typically be categorized into three main classes based on their stage of intervention relative to the server's aggregation step:  pre-aggregation detection \cite{DBLP:journals/access/YeoL25,DBLP:journals/access/MasudaKKTH23,DBLP:journals/access/GhaziFN25,DBLP:journals/access/Almuseelem25}, robust aggregation \cite{DBLP:journals/tsp/PillutlaKH22,DBLP:conf/icml/YinCRB18,DBLP:conf/aaai/OzdayiKG21}, and post-aggregation mitigation \cite{DBLP:journals/corr/abs-1911-07963,DBLP:journals/corr/abs-2011-01767}. 

The core idea of post-aggregation Mitigation \cite{DBLP:journals/corr/abs-1911-07963,DBLP:journals/corr/abs-2011-01767} is to purify or repair the global model after accepting and aggregating all client gradients, in order to alleviate the negative impact of backdoors.
Weak-DP \cite{DBLP:journals/corr/abs-1911-07963} directly applies the Differential Privacy (DP) \cite{DBLP:journals/fttcs/DworkR14}, mitigating malicious effects by disturbing potential backdoor patterns through norm clipping and the injection of Gaussian noise into the global model. 
Pruning \cite{DBLP:journals/corr/abs-2011-01767} is based on the assumption that the backdoor task and the main task activate different neurons. Therefore, this method attempts to identify and prune these neurons believed to be associated with the backdoor. 
Although these defenses are effective to some extent, their common limitation is significant: in the process of eliminating backdoors, they often unavoidably damage the global model's performance on the main task. 

The goal of robust aggregation \cite{DBLP:journals/tsp/PillutlaKH22,DBLP:conf/icml/YinCRB18,DBLP:conf/aaai/OzdayiKG21,DBLP:journals/tifs/ZhangZSGCSY24} is to design aggregation rules that are inherently resilient to malicious gradients. Such defenses do not attempt to explicitly identify which clients are malicious, but rather aim to directly weaken the influence of malicious gradients on the global model during the aggregation process through robust statistical or filtering mechanisms.
RFA \cite{DBLP:journals/tsp/PillutlaKH22} employs the Geometric Median of client gradients to compute the global gradient, which is a statistically more robust defense.
Trimmed-Mean \cite{DBLP:conf/icml/YinCRB18} removes a fraction of the largest and smallest values from all client updates before aggregation and then averages only the remaining central values. 
RLR \cite{DBLP:conf/aaai/OzdayiKG21} adopts a dynamic adjustment approach by analyzing the sign of each client gradient on a per-dimension basis and adjusting the global learning rate for that dimension accordingly. This defense aims to corrupt the specific gradient direction required for backdoor injection and is combined with noise to enhance its effect.
Although these defenses enhance the robustness of the FL process, due to the stealthy nature of backdoor attacks, they cannot completely eliminate the malicious impact. 

The core principle of pre-aggregation detection \cite{DBLP:conf/nips/BlanchardMGS17,DBLP:conf/ndss/CaoF0G21,DBLP:conf/iccv/HuangLCS023,DBLP:conf/raid/FungYB20,DBLP:journals/tifs/MaMMLD22,DBLP:conf/uss/NguyenRCYMFMMMZ22,DBLP:journals/tdsc/MuCSLCZM24} is to directly identify and filter out malicious gradients by analyzing client-uploaded gradients prior to aggregation. 
Multi-Krum \cite{DBLP:conf/nips/BlanchardMGS17}, an extension of the Krum, first calculates a score for each client's gradient, defined as the sum of its squared Euclidean distances to its nearest neighbors. It then selects the gradients with the lowest scores, and their average is used as the global gradient.  
FLtrust \cite{DBLP:conf/ndss/CaoF0G21} requires the server to maintain a clean dataset and a root model to validate client gradients via cosine similarity. However, this requirement for a clean dataset on the server contradicts the privacy-preserving nature of FL, making it difficult to deploy in practice. 
To more flexibly identify malicious clients, many modern defenses have turned to clustering detection. 
Multi-metrics \cite{DBLP:conf/iccv/HuangLCS023} uses multiple distance metrics, such as Manhattan, Euclidean and cosine, and introduces a dynamic weighting mechanism, ultimately aggregating only a small subset of clients deemed most trustworthy. 
Scope \cite{DBLP:journals/tifs/HuangLYGCSCN25} does not rely on multi-metrics. Instead, it operates by transforming the gradients themselves, aiming to fundamentally amplify the hidden backdoor features within malicious gradients to more effectively counter stealthy backdoor attacks. 

\section{System and Problem Setting}
\label{sec: system}
\subsection{System Setting}
In our FL system, at the beginning of each training round $t$, the server randomly selects $k$ clients from all clients to participate in the current round. We assume that among these $k$ selected clients, there are $c$ malicious clients controlled by an attacker and $k-c$ honest clients. We assume that honest clients constitute the majority in the selected subset, i.e., $k-c > k/2$. Subsequently, the server receives the gradients $\{g_1^t, \ldots, g_k^t\}$ from these $k$ clients. By default, the server utilizes the Federated Averaging (FedAvg) \cite{DBLP:conf/aistats/McMahanMRHA17} algorithm to aggregate these gradients and generate the new global model $g^t$.

\subsection{Threat Model}
We consider an attacker acting as a malicious client in the FL process, whose objective is to inject a backdoor into the final global model $g$ by manipulating its local gradients. A successful backdoor attack must satisfy two conditions: effectiveness, where the final global model is forced to output a pre-defined target label when encountering inputs embedded with a specific trigger, and stealthiness, where the model's performance on the main task with benign (clean) inputs remains indistinguishable from that of an honest model, thereby evading detection. We assume a partial-knowledge \cite{DBLP:conf/uss/FangCJG20} setting where the attacker has full control over their local training process, including the ability to arbitrarily modify their dataset and loss function. Furthermore, the attacker is aware of the aggregation rule, such as FedAvg. However, these capabilities are strictly limited: the attacker cannot access the data of other benign clients, nor can they interfere with the server's aggregation procedure or the training routines of other honest participants. All other assumptions adhere to the framework established in Scope \cite{DBLP:journals/tifs/HuangLYGCSCN25}. 

\subsection{Defense Goals}
To effectively counter sophisticated backdoor attacks, we enhance the existing Scope by proposing a novel defense. Our method is designed to achieve the following four key objectives:

\begin{enumerate}
    \item \textbf{Effectiveness:} As the primary objective, the FAROS should accurately identify and filter out backdoored models uploaded by malicious clients. In the face of diverse backdoor attacks, it should maintain the Attack Success Rate (ASR) of the global model at a minimal level.

    \item \textbf{Accuracy:} While ensuring defensive effectiveness, the method should not compromise the model's benign performance on the main task. In a non-attack setting, the accuracy of the global model trained with our method should be on par with that of the standard FedAvg.

    \item \textbf{Efficiency:} To ensure its practicality, the FAROS should introduce minimal additional computational overhead. The overall training time should be comparable to that of the standard FedAvg.

    \item \textbf{Generalizability:} The FAROS should not be confined to a specific FL setting. It must maintain robust performance across a variety of settings, including but not limited to a varying number of clients, different degrees of non-iid, and various benchmark datasets.
\end{enumerate}

\section{Methodology}
\label{sec: method}

\subsection{Overview}
This paper proposes FAROS, a more robust and adaptive defense. 
The scheme employs a two-stage dynamic abnormal gradient detection process to effectively detect gradients prior to aggregation. 
The first component of FAROS is ADS. 
Instead of using fixed parameters, this component first analyzes the statistical properties of all client gradients in the current round to infer potential attack patterns. 
Subsequently, it adaptively adjusts the degree of scaling based on this assessment to more precisely amplify the anomalous values within suspicious gradients. 
The second component is RCC. 
RCC addresses the vulnerability of traditional defenses, which can be sensitive to the choice of a single starting point for clustering.
It establishes a stable benchmark by identifying a robust core set composed of the most trustworthy clients. 
Ultimately, all gradients are partitioned based on their similarity to this benchmark, enabling the reliable identification and exclusion of suspicious malicious gradients from the aggregation process. 
Fig. \ref{sec: method} illustrates the overall framework of FAROS. 

\begin{figure}[!t]
\centering
\includegraphics[width=\textwidth]{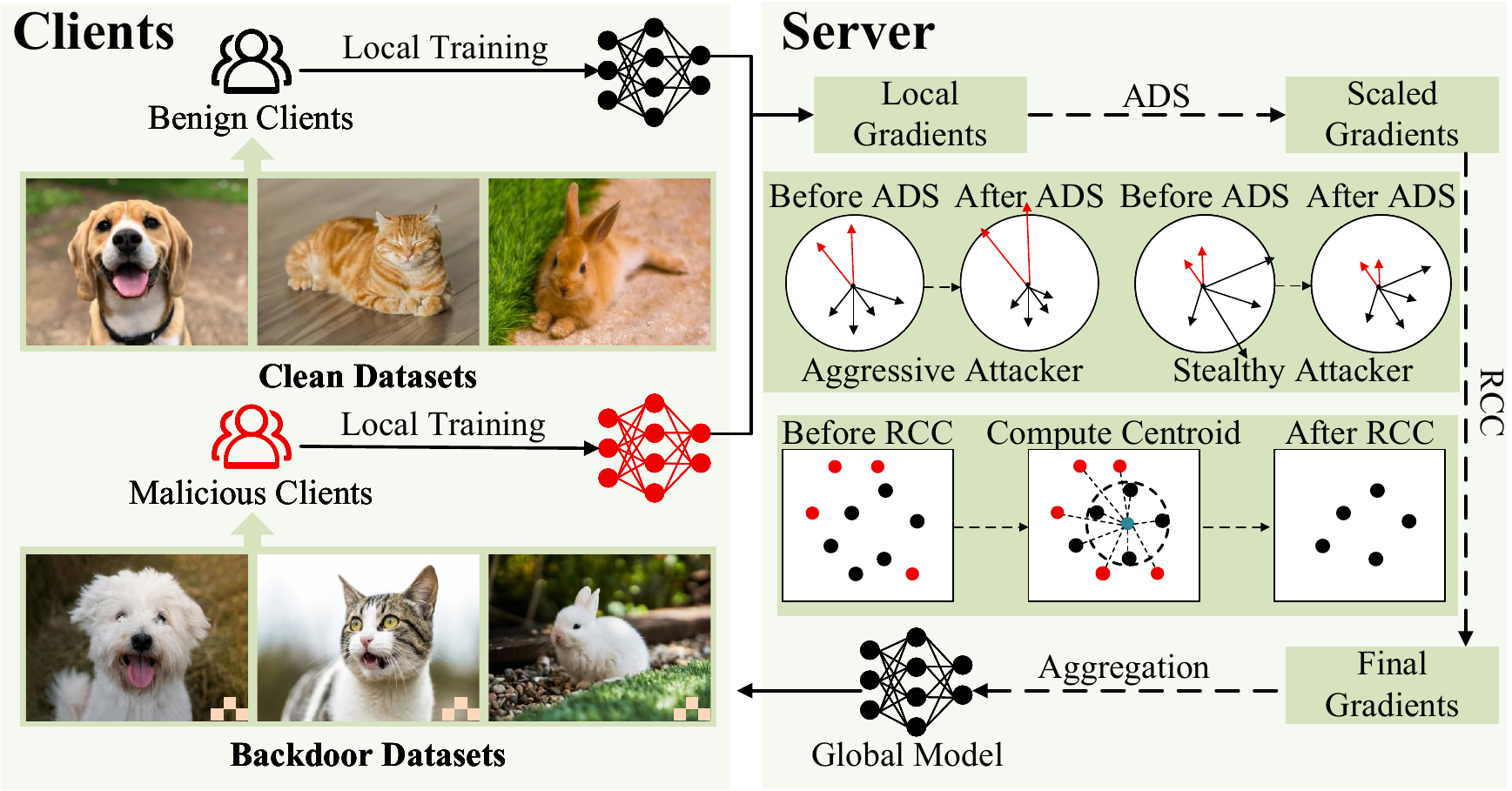}
\caption{Overview of FAROS.}
\label{overview}
\end{figure}

\subsection{Adaptive Differential Scaling}
Our adaptive differential scaling amplifies the dimensions of the normalized gradient that exhibit significant changes, using a fixed power function:
\begin{equation}
\label{eq:static_scaling}
(g_i^j)^* = \left( |g_i^{j}| \right)^{\varphi} \cdot \text{sgn}(g_i^{j})
\end{equation}
Where $g_i^j$ denotes the $j$-th dimension of the $i$-th client's gradient, and the scaling factor $\varphi$ is a parameter that must be set in advance. However, the imitation of this static strategy is that it employs a fixed pattern against a dynamic attacker.
A sophisticated attacker can trade off between attack effectiveness and stealthiness via the loss function:
\begin{equation}
\mathcal{L}_{\textit{attack}} = (1 - \alpha) \mathcal{L}_{\textit{class}} + \alpha \mathcal{L}_{\textit{cosine}}
\label{eq:attack_loss}
\end{equation}
where the hyperparameter $\alpha \in [0, 1]$ balances the classification loss (effectiveness) against the cosine distance loss (stealthiness). A smaller $\alpha$ leads to more aggressive gradients, whereas a larger $\alpha$ results in stealthier ones. Therefore, a fixed scaling factor $\varphi$ is unlikely to achieve optimal performance across the entire spectrum of attack strategies determined by $\alpha$. 

To construct a defense that is robust against the entire spectrum of attack strategies, we assume that the attack style can be inferred from the overall distribution of client gradients. In each round $t$, the server computes the dispersion of the dimension-wise normalized gradients from all uploaded clients, $\{g_1^*, \ldots, g_k^*\}$. We first compute the centroid of all vectors:
\begin{equation}
g_c = \frac{1}{K} \sum_{i=1}^K g_i^*
\end{equation}
Subsequently, the variance of the cosine distances from all gradients to this centroid is calculated, which serves as the current round's dispersion, $D_t$:
\begin{equation}
D_t = \text{Var}\left( \left\{ 
1 - \frac{\langle g_i^*, g_c \rangle}{\| g_i^* \| \cdot \| g_c \|}
\right\}_{i=1}^K \right)
\end{equation}
The value of $D_t$ effectively reflects the degree of inconsistency among the gradients in the current round. A small $D_t$ value typically signifies the presence of extremely stealthy attacks, whereas a large $D_t$ value indicates more overt attacks. 

The scaling factor $\varphi_t$ for the current round is dynamically computed based on the dispersion metric $D_t$. We employ an exponential decay function to establish a relationship between $\varphi_t$ and $D_t$:
\begin{equation}
\varphi_t = 1 + (\varphi_{\max} - 1) \cdot \exp(-k \cdot D_t)
\label{eq:adaptive_phi}
\end{equation}
Where $\varphi_{\max}$ is the preset maximum scaling factor, and $k$ is a positive parameter that controls the rate of decay. This formulation ensures that when $D_t$ is small, $\varphi_t$ approaches $\varphi_{\max}$, leading to a strong scaling effect that enhances the capability of slight differences. Conversely, when $D_t$ is large, $\varphi_t$ approaches 1, resulting in a moderate scaling to maintain the robustness of the defense. Finally, we perform ADS on each gradient. Specifically, this is achieved by replacing the fixed $\varphi$ in the formula \ref{eq:static_scaling} with the dynamically computed $\varphi_t$.

\subsection{Robust Core-set Computing}

Our RCC robustly identifies benign clients in a three-step process. Its core principle is to replace the uncertain single dominant gradient with a stable centroid. 
First, RCC selects the $l$ clients with the minimum distances to form a robust core set, $S_{core}$. This core set represents the group of clients with the highest confidence in the current round.
After obtaining the set, RCC computes a centroid ($\mathbf{g}_{centroid}$) by averaging all gradients within this set:
\begin{equation}
\mathbf{g}_{centroid} = \frac{1}{l} \sum_{j \in S_{core}} \mathbf{g}_j^*
\end{equation}
This centroid can be regarded as the most representative benign gradient of the current round. Compared to any single gradient, $\mathbf{g}_{centroid}$ is an averaged value, making it more stable and less susceptible to the influence of outliers. 
With this benchmark established, RCC then calculates the cosine distance between the gradients of all $k$ clients and this centroid:
\begin{equation}
d_i = 1 - \frac{\langle \mathbf{g}_i^*, \mathbf{g}_{centroid} \rangle}{\|\mathbf{g}_i^*\| \cdot \|\mathbf{g}_{centroid}\|}
\end{equation}
Finally, the clients with the shortest distance to this centroid are selected as the final set of benign clients. 

\begin{algorithm}[t!]
\small
\LinesNumbered
\caption{FAROS Algorithm}
\label{alg:ares}
\KwIn{
    Current round's client models $\{g_1^t, \ldots, g_k^t\}$, previous round's global model $g^{t-1}$, max scaling factor $\varphi_{\max}$, core-set size $l$, Global learning rate $\eta$.
} 
\KwOut{New global model $g^t$.}

\textbf{Server performs:} \\
\For{$t = 1, 2, \ldots, T$}{
    \tcp{Step 1: Normalization}
    \For{$i = 1$ \textbf{to} $k$}{
        $\mathbf{g}_i^* \gets \text{Normalize}(g_i^t - g^{t-1})$ 
    }
    
    \tcp{Step 2: ADS}
    $\mathbf{g}_c \gets \frac{1}{k} \sum_{i=1}^k \mathbf{g}_i^*$
    
    $D_t \gets \operatorname{Var}\left( \left\{ 1 - \frac{\langle \mathbf{g}_i^*, \mathbf{g}_c \rangle}{\| \mathbf{g}_i^* \| \cdot \| \mathbf{g}_c \|} \right\}_{i=1}^k \right)$
    
    $\varphi_t \gets 1 + (\varphi_{\max} - 1) \cdot \exp(-\kappa \cdot D_t)$
    
    \For{$i = 1$ \textbf{to} $k$}{
        \For{each dimension $j$}{
            $(g_i^j)^* = \left( |g_i^{j}| \right)^{\varphi_t} \cdot \text{sgn}(g_i^{j})$
        }
    }
    
    \tcp{Step 3: RCC}
    \For{$i = 1$ \textbf{to} $k$}{
        $\delta_i \gets \sum_{p=1}^{k} \left(1 - \frac{\langle \mathbf{g}_i^{*}, \mathbf{g}_p^{*} \rangle}{\|\mathbf{g}_i^{*}\| \cdot \|\mathbf{g}_p^{*}\|}\right)$ 
    }
    $S_{\text{core}} \gets \text{top-$l$ clients with minimal } \delta_i$ 
    
    $\mathbf{g}_{\text{centroid}} \gets \frac{1}{l} \sum_{j \in S_{\text{core}}} \mathbf{g}_j^{*}$
    
    \For{$i = 1$ \textbf{to} $k$}{
        $d_i \gets 1 - \frac{\langle \mathbf{g}_i^{*}, \mathbf{g}_{\text{centroid}} \rangle}{\|\mathbf{g}_i^{*}\| \cdot \|\mathbf{g}_{\text{centroid}}\|}$ 
    }
    $C_{\text{benign}} \gets \text{top-$m$ clients with minimal } d_i$ 
    
    \tcp{Step 4: Aggregation}
    $\mathbf{g}_{\text{agg}} \gets \frac{1}{m} \sum_{i \in C_{\text{benign}}} (g_i^t - g^{t-1})$
    
    $g^t \gets g^{t-1} + \eta \cdot \mathbf{g}_{\text{agg}}$
}
\Return{$g^t$}
\end{algorithm}

\subsection{Complete FAROS Algorithm}
Algorithm  \ref{alg:ares} details the process of our FAROS. In each round $t$, every client receives the global model $g^{t-1}$ from the previous round, computes its local model $g_i^t$ based on its local data, and sends the gradients to the server. The server then processes the gradients from $k$ randomly selected clients as follows: first, it normalizes each gradient difference to obtain the normalized gradient $\mathbf{g}_i^*$. Second, the server computes the centroid of the normalized gradients, $\mathbf{g}_c$, and calculates a dispersion $D_t$ based on the variance of the cosine distances between each $\mathbf{g}_i^*$ and $\mathbf{g}_c$. It then generates a dynamic scaling factor $\varphi_t$ to perform dimension-wise scaling on $\mathbf{g}_i^*$, yielding the scaled gradient. Third, the algorithm selects a core set $S_{\text{core}}$ based on the sum of pairwise cosine distances $\delta_i$ for each scaled gradient $\mathbf{g}_i^{*}$. After computing this set's centroid $\mathbf{g}_{\text{centroid}}$, the algorithm filters for $m$ benign clients $C_{\text{benign}}$ based on the cosine distance $d_i$ of each gradient to this centroid. Finally, the server averages the raw gradient differences from the clients in $C_{\text{benign}}$ to obtain the aggregated gradient, $\mathbf{g}_{\text{agg}}$, and updates the global model with a learning rate $\eta$. This process iterates until the global model converges.

\section{Experimental Evaluation}
\label{sec: result}
\subsection{Experiment Setup}

\subsubsection{System Settings}
Our experimental evaluation was conducted on two benchmark datasets. 
\begin{itemize}
    \item \textbf{CIFAR10 \cite{krizhevsky2009learning}} is a classic color image dataset extensively used for object recognition research. It comprises 60,000 low-resolution color images of size $32 \times 32$ pixels, evenly distributed across 10 mutually exclusive classes (e.g., airplane, automobile, bird, etc.). The dataset is partitioned into a training set of 50,000 images and a test set of 10,000 images.

    \item \textbf{EMNIST \cite{DBLP:conf/ijcnn/CohenATS17}} is an extension of the classic MNIST. To align our experiments with classic digit recognition tasks, we utilized the Digits subset. This subset is exclusively composed of handwritten digits (0-9) across 10 classes, with all images being $28 \times 28$ pixel grayscale images. 
\end{itemize}
For model architectures, we followed the standard configurations from \cite{DBLP:journals/tifs/HuangLYGCSCN25}, selecting models appropriate for the complexity of each dataset. Specifically, we employed the VGG-9 \cite{DBLP:journals/corr/SimonyanZ14a} for CIFAR10, whereas for the simpler EMNIST, we utilized the classic LeNet-5 \cite{DBLP:journals/pieee/LeCunBBH98}, which serves as an efficient benchmark. 
Our FL setup simulates 200 total clients, from which 20 are randomly sampled in each communication round to participate in local training and server aggregation. We set the total number of training rounds to 500 for the CIFAR10 and 300 for the EMNIST. Our FL setup simulates 200 total clients, from which 20 are randomly sampled in each communication round to participate in local training and server aggregation. We set the total number of training rounds to 500 for the CIFAR10 and 300 for the EMNIST.
To simulate a non-iiD scenario, we partitioned the dataset using a Dirichlet distribution with a concentration parameter of $q=0.4$. In this framework, a lower q value corresponds to a higher degree of data heterogeneity among clients.

\subsubsection{Compared Attacks and Defenses}We evaluate the effectiveness of our FAROS using three representative backdoor attacks: Model Replacement Attack \cite{DBLP:conf/aistats/BagdasaryanVHES20}, Constrain-and-scale Attack \cite{DBLP:conf/aistats/BagdasaryanVHES20}, and Edge-case PGD Attack \cite{DBLP:conf/nips/WangSRVASLP20} for which the implementation details and parameter settings are adopted from Scope \cite{DBLP:journals/tifs/HuangLYGCSCN25}. 

We employ the standard FedAvg \cite{DBLP:conf/aistats/McMahanMRHA17} as the performance baseline. To validate the defensive capabilities of our FAROS, we conduct a comprehensive comparison against three state-of-the-art defenses, namely Multi-Krum \cite{DBLP:conf/nips/BlanchardMGS17}, Weak-DP \cite{DBLP:journals/corr/abs-1911-07963}, and Scope \cite{DBLP:journals/tifs/HuangLYGCSCN25}. These methods were selected because they cover prevailing technical approaches in FL, thus providing robust comparative benchmarks. All experimental parameters not explicitly mentioned in this section adhere to the settings in Scope \cite{DBLP:journals/tifs/HuangLYGCSCN25}. 

\subsubsection{Evaluation Metrics}
We assess the effectiveness of defenses using two primary metrics: Attack Success Rate (ASR) and Main task Accuracy (ACC). ASR quantifies the success rate on the backdoor task; it is a metric the attacker seeks to maximize and the defender seeks to minimize. ACC represents the model's performance on its original benign task. A robust defense must maintain a high ACC, ensuring that the defense mechanism itself does not significantly degrade the model's utility. 

\subsection{Experimental Results}
\begin{table*}[ht!]
\centering
\caption{Robustness against Different Attacks and Defenses. }
\label{tab:defense_comparison_partial}
\resizebox{\textwidth}{!}{%
\begin{tabular}{@{}llcccccc@{}}
\toprule
\multirow{2}{*}{\textbf{Dataset}} & \multirow{2}{*}{\textbf{Defense}} & \multicolumn{2}{c}{\textbf{Model Replacement \cite{DBLP:conf/aistats/BagdasaryanVHES20}}} & \multicolumn{2}{c}{\textbf{Constrain-and-scale \cite{DBLP:conf/aistats/BagdasaryanVHES20}}} & \multicolumn{2}{c}{\textbf{Edge-case PGD \cite{DBLP:conf/nips/WangSRVASLP20}}} \\
\cmidrule(lr){3-4} \cmidrule(lr){5-6} \cmidrule(lr){7-8}
& & ACC $\uparrow$ & ASR $\downarrow$ & ACC $\uparrow$ & ASR $\downarrow$ & ACC $\uparrow$ & ASR $\downarrow$ \\
\midrule
\multirow{6}{*}{\hspace{1.5em}\rotatebox{90}{CIFAR10 \cite{krizhevsky2009learning}}}
& FedAvg \cite{DBLP:conf/aistats/McMahanMRHA17}      & 85.21 & 65.21 & 87.37 & 13.88 & 88.17 & 63.65 \\
& Multi-krum \cite{DBLP:conf/nips/BlanchardMGS17} & 83.21 & 2.87 & 86.54 & 26.76 & 85.91 & 56.71 \\
& Weak-DP \cite{DBLP:journals/corr/abs-1911-07963}    & 81.34 & 19.11 & 79.34 & 24.64 & 72.11 & 15.75 \\
& Scope \cite{DBLP:journals/tifs/HuangLYGCSCN25}      & 84.78 & 1.62 & 86.32 & 4.56 & 85.12 & 5.12 \\
\cmidrule(lr){2-8}
& FAROS & 85.14 & 0.52 & 86.33 & 3.43 & 85.87 & 2.67 \\
\midrule
\multirow{6}{*}{\hspace{1.5em}\rotatebox{90}{EMNIST \cite{DBLP:conf/ijcnn/CohenATS17}}}
& FedAvg \cite{DBLP:conf/aistats/McMahanMRHA17}      & 99.12 & 95.67 & 98.56 & 53.23 & 99.11 & 88.37 \\
& Multi-krum \cite{DBLP:conf/nips/BlanchardMGS17} & 98.26 & 35.89 & 98.33 & 14.18 & 97.41 & 29.97 \\
& Weak-DP \cite{DBLP:journals/corr/abs-1911-07963}    & 98.68 & 9.21 & 98.45 & 51.45 & 98.65 & 85.02 \\
& Scope \cite{DBLP:journals/tifs/HuangLYGCSCN25}      & 98.89 & 5.74 & 98.49 & 4.84 & 98.41 & 16.12 \\
\cmidrule(lr){2-8}
& FAROS & 98.48 & 1.82 & 98.43 & 2.16 & 98.83 & 3.78 \\
\bottomrule
\end{tabular}%
}
\end{table*} 
The experiments comprehensively evaluate our FAROS to validate its achievement of the defense goals defined in Section \ref{sec: system}. 
\begin{itemize}
    \item \textbf{Effectiveness \& Accuracy:} We demonstrate through the comprehensive performance comparison in Table \ref{tab:defense_comparison_partial} that FAROS effectively defends against various backdoor attacks while maintaining main task accuracy.

    \item \textbf{Generalizability:} The results in Fig. \ref{fig:non-iid} and Fig. \ref{fig:number of clients} validate our method's generalizability by demonstrating its robust performance across various FL settings.

    \item \textbf{Efficiency:} The runtime analysis in Table \ref{tab:runtime_comparison_enhanced} confirms that our FAROS achieves its performance improvements without introducing significant computational overhead, thereby satisfying the efficiency.
    
    \item \textbf{Ablation Study:} We conduct ablation studies to independently validate the effectiveness of our two core components. The results, presented in Table \ref{tab:detailed_component_comparison}, demonstrate that their combination achieves optimal performance.
\end{itemize}

\subsubsection{Robustness against Different Attacks and Defenses}
The comprehensive performance comparison in Table \ref{tab:defense_comparison_partial} clearly demonstrates the superiority of our FAROS. The results first reveal the severe shortcomings of traditional defenses like Multi-krum and Weak-DP against sophisticated attacks. Notably, on the CIFAR10 against the Constrain-and-scale attack, Multi-krum records an ASR of 46.76\%, a figure substantially higher than the FedAvg (13.88\%). This indicates that its erroneous gradient selection mechanism can have a detrimental effect under such conditions. 
Our method not only surpasses these traditional schemes but also achieves key performance improvements over Scope. On CIFAR10, our method demonstrates a comprehensive advantage, particularly against the Edge-case PGD attack. It not only further reduces the ASR from 5.12\% to 4.67\% but also improves the ACC from 85.12\% to 85.87\%, achieving a Pareto improvement in security and performance. This advantage is also evident on the EMNIST. It still achieves a lower ASR (5.78\% vs. 6.12\%) against the Edge-case PGD attack, further validating its robustness. In summary, the experimental results provide strong evidence that our method strikes an optimal balance between Effectiveness and Accuracy, successfully achieving its objectives.

\subsubsection{Impact of the Degree of Non-IID}
\begin{figure}[!htbp]
\centering
\includegraphics[width=\textwidth]{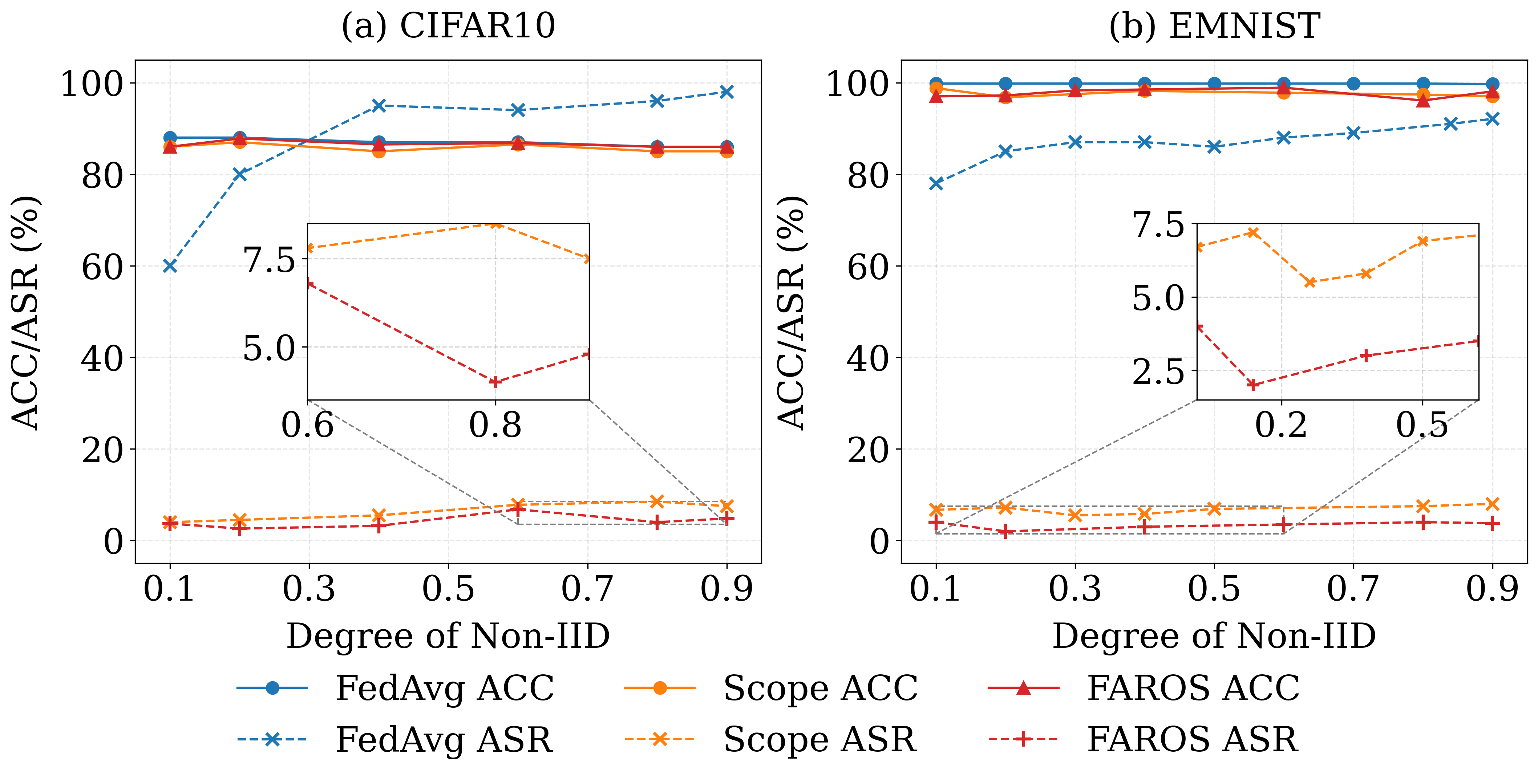}
\caption{Impact of the Degree of Non-IID.}
\label{fig:non-iid}
\end{figure}

Fig. \ref{fig:non-iid} illustrates the performance comparison of our FAROS, Scope, and FedAvg on CIFAR10 and EMNIST under varying degrees of non-iid. In high non-iid scenarios on CIFAR10, our method exhibits stronger backdoor robustness than Scope, consistently achieving a lower ASR. We attribute this enhanced robustness to our RCC algorithm, which uses a consensus core-set mechanism to more accurately identify benign clients in highly non-iid environments, thus avoiding the erroneous filtering caused by data distribution shifts. On the EMNIST, our method demonstrates higher defense precision in lower non-iid environments (e.g., achieving an ASR of only 3\% compared to Scope's nearly 6\% at a non-iiD degree of 0.4), while maintaining on par with Scope in high non-iid settings. In summary, the experiments demonstrate that our method is not only more robust in extreme scenarios but also maintains excellent performance in regular settings. 

\subsubsection{Impact of the Total Number of Clients}
\begin{figure}[!htbp]
\centering
\includegraphics[width=\textwidth]{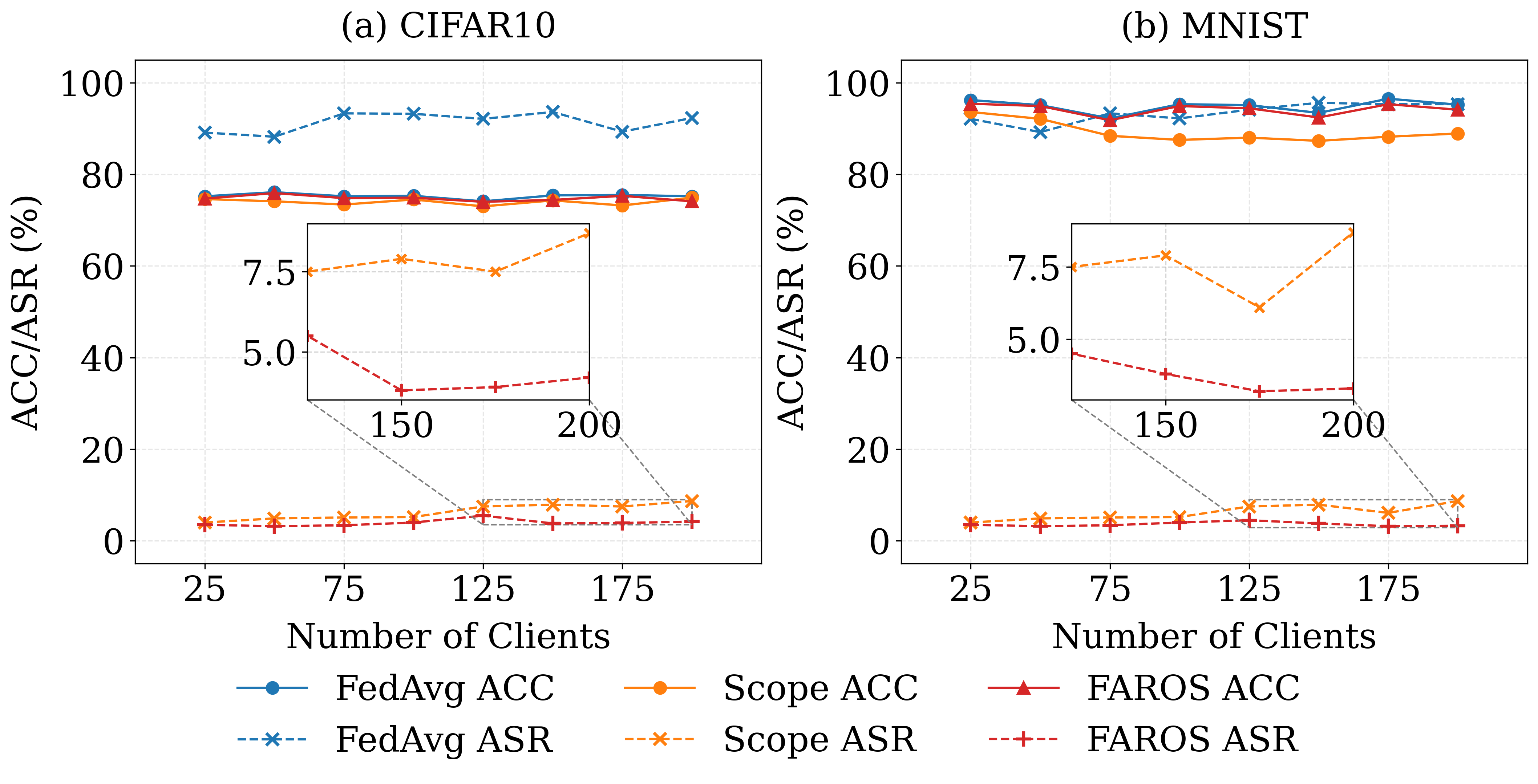}
\caption{Impact of the Total Number of Clients.}
\label{fig:number of clients}
\end{figure}

Fig. \ref{fig:number of clients} presents a performance comparison of our FAROS, Scope, and FedAvg on the CIFAR10 and EMNIST as the total number of clients varies. On the CIFAR10, our method demonstrates excellent stability. Notably, as the number of clients increases to 150-175, Scope's defense performance fluctuates sharply with its ASR peaking at 8.0\%. In contrast, our method successfully suppresses this peak to within 6.0\%, effectively avoiding a sharp degradation in performance. This advantage behaves as consistent performance leadership on the EMNIST. For instance, at 100 clients, our method's ASR (3.8\%) is significantly lower than Scope's (5.8\%). When representing a 40\% improvement in defense effectiveness compared to Scope's 5.0\%.  
Collectively, the results from Fig. \ref{fig:non-iid} and Fig. \ref{fig:number of clients} demonstrate that our method achieves the goal of generalizability. 

\subsubsection{Ablation Study}
\begin{table}[ht]
\centering
\caption{Ablation Study. }
\label{tab:detailed_component_comparison}
\resizebox{0.8\textwidth}{!}{
\begin{tabular}{lcccc} 
\toprule
\multirow{2}{*}{\textbf{Component}} & \multicolumn{2}{c}{\textbf{Model Replacement}} & \multicolumn{2}{c}{\textbf{Edge-case PGD}} \\ 
\cmidrule(lr){2-3} \cmidrule(lr){4-5} 
& ACC $\uparrow$ & ASR $\downarrow$ & ACC $\downarrow$ & ASR $\uparrow$ \\ 
\midrule
Scope                       & 84.86 & 0.71 & 84.32 & 6.02 \\ 
Only ADS                    & 84.75 & 0.73 & 83.21  & 8.23  \\
Only RCC                    & 84.85 & 0.66 & 83.56 &  6.94 \\
FAROS   & 84.56 & 0.65 & 83.98 & 4.42 \\
\bottomrule
\end{tabular}
}
\end{table}

We conducted ablation studies on the CIFAR-10 to validate the contribution of each component within our FAROS. As detailed in Table \ref{tab:detailed_component_comparison}.
When the RCC component was employed (Only RCC), it improved defense against Model Replacement, reducing the ASR from 0.71 to 0.66. However, it performed poorly against Edge-case PGD, underperforming the Scope. The ADS component was inferior to the Scope across all metrics in both attack scenarios, indicating the limited defensive capability of a single component. 
In contrast, our FAROS demonstrated a stronger defensive performance. Under the Model Replacement, it achieved the lowest ASR of 0.65, at the cost of only a marginal decrease in ACC. The superiority of FAROS was particularly pronounced under the Edge-case PGD, where it reduced the ASR to 5.42, significantly outperforming the Scope (6.02) and all other components. 
In summary, our ablation study strongly demonstrates the synergy between the ADS and RCC components. Although the individual components are of limited use on their own, their integration leads to a robust framework.

\subsubsection{Computational Cost}
\begin{table}[ht]
\centering
\caption{Comparison of Runtime.}
\resizebox{0.4\textwidth}{!}{
\label{tab:runtime_comparison_enhanced}
\renewcommand{\arraystretch}{1.2}
\begin{tabular}{cc}
\toprule
\textbf{Defense} &  \textbf{Runtime}  \\ 
\midrule
FedAvg      & 25.32 \\
Multi-krum  & 29.28 \\
Weak-DP     & 26.86 \\
Scope       & 28.78 \\
FAROS       & 26.18 \\ 
\bottomrule
\end{tabular}
}
\end{table}

To evaluate the computational overhead of our FAROS, we measured and compared its runtime against several baseline defenses on the EMNIST. All experiments were conducted within a consistent experimental environment to ensure a fair comparison. The results are presented in Table \ref{tab:runtime_comparison_enhanced}.
The standard FedAvg, which lacks additional defenses, exhibited the shortest runtime (25.32s) and thus serves as a performance baseline for measuring computational overhead. Our FAROS recorded a runtime of 28.89s, which introduces an overhead of approximately 14.1\% relative to the FedAvg. When compared with other defenses, the computational cost of our method is comparable to that of Multi-krum (29.28s), the most computationally intensive defense, and slightly higher than Weak-DP (26.86s) and Scope (27.78s). 
Despite this additional computational overhead, we argue that it is entirely acceptable and represents a favorable trade-off. As demonstrated in Table \ref{tab:defense_comparison_partial}, our method is significantly superior to Weak-DP and Scope in terms of robustness and performance. Therefore, this increase in runtime is necessary and worthwhile for a substantial improvement in model security. 

\section{Conclusion}
\label{sec: conclusion}
This paper introduces FAROS, an enhanced defense mechanism against backdoor attacks in FL. 
Specifically, the framework utilizes an innovative ADS mechanism to dynamically adjust its defense sensitivity based on the dispersion of client gradients in each round. Following this, client similarity is evaluated using our designed RCC algorithm. 
The RCC algorithm effectively identifies and filters out malicious clients by computing around the centroid of a high-confidence core set, thus significantly reducing the risk of single-point failures.
Extensive experiments conducted across various datasets, models, and attack scenarios demonstrate that our method outperforms state-of-the-art defenses, achieving lower rates of defense success, improved task accuracy, and greater generalizability.

In future work, we plan to extend this research in the following directions:
(i) Optimizing Components. We aim to explore more advanced dynamic adjustment strategies for the ADS. This may include the introduction of Reinforcement Learning (RL) to determine the optimal scaling factor for each round automatically.
(ii) Integration with Privacy Technologies. We will investigate how to combine our robustness mechanisms with techniques such as Differential Privacy (DP) and Homomorphic Encryption (HE). Our goal is to build a comprehensive security framework for FL that can simultaneously defend against both poisoning attacks and inference attacks.

\section*{Acknowledgement}
This research was partially supported by JSPS KAKENHI (No. 25K15290). This work was partially conducted during the internship at Waseda University.

%
%
%
\bibliographystyle{splncs04}
\bibliography{bib}

@article{krizhevsky2009learning,
  title={Learning multiple layers of features from tiny images},
  author={Krizhevsky, Alex and Hinton, Geoffrey and others},
  year={2009}
}

@inproceedings{DBLP:conf/ijcnn/CohenATS17,
  author       = {Gregory Cohen and
                  Saeed Afshar and
                  Jonathan Tapson and
                  Andr{\'{e}} van Schaik},
  title        = {{EMNIST:} Extending {MNIST} to handwritten letters},
  booktitle    = {International Joint Conference on Neural Networks},
  pages        = {2921--2926},
  year         = {2017}
}

@article{DBLP:journals/tifs/HuangLYGCSCN25,
  author       = {Siquan Huang and
                  Yijiang Li and
                  Xingfu Yan and
                  Ying Gao and
                  Chong Chen and
                  Leyu Shi and
                  Biao Chen and
                  Wing W. Y. Ng},
  title        = {Scope: On Detecting Constrained Backdoor Attacks in Federated Learning},
  journal      = {IEEE Transactions on Information Forensics and Security
  },
  volume       = {20},
  pages        = {3302--3315},
  year         = {2025}
}

@inproceedings{DBLP:journals/corr/SimonyanZ14a,
  author       = {Karen Simonyan and
                  Andrew Zisserman},
  title        = {Very Deep Convolutional Networks for Large-Scale Image Recognition},
  booktitle    = {International Conference on Learning Representations},
  year         = {2015}
}

@article{DBLP:journals/pieee/LeCunBBH98,
  author       = {Yann LeCun and
                  L{\'{e}}on Bottou and
                  Yoshua Bengio and
                  Patrick Haffner},
  title        = {Gradient-based learning applied to document recognition},
  journal      = {Proceedings of the IEEE},
  volume       = {86},
  number       = {11},
  pages        = {2278--2324},
  year         = {1998}
}

@inproceedings{DBLP:conf/aistats/McMahanMRHA17,
  author       = {Brendan McMahan and
                  Eider Moore and
                  Daniel Ramage and
                  Seth Hampson and
                  Blaise Ag{\"{u}}era y Arcas},
  title        = {Communication-Efficient Learning of Deep Networks from Decentralized
                  Data},
  booktitle    = {International Conference on Artificial Intelligence and Statistics},
  volume       = {54},
  pages        = {1273--1282},
  year         = {2017}
}

@inproceedings{DBLP:conf/nips/BlanchardMGS17,
  author       = {Peva Blanchard and
                  El Mahdi El Mhamdi and
                  Rachid Guerraoui and
                  Julien Stainer},
  title        = {Machine Learning with Adversaries: Byzantine Tolerant Gradient Descent},
  booktitle    = {Neural Information Processing Systems},
  pages        = {119--129},
  year         = {2017}
}

@inproceedings{DBLP:conf/aistats/BagdasaryanVHES20,
  author       = {Eugene Bagdasaryan and
                  Andreas Veit and
                  Yiqing Hua and
                  Deborah Estrin and
                  Vitaly Shmatikov},
  title        = {How To Backdoor Federated Learning},
  booktitle    = {International Conference on Artificial Intelligence and Statistics},
  volume       = {108},
  pages        = {2938--2948},
  year         = {2020}
}

@inproceedings{DBLP:conf/nips/WangSRVASLP20,
  author       = {Hongyi Wang and
                  Kartik Sreenivasan and
                  Shashank Rajput and
                  Harit Vishwakarma and
                  Saurabh Agarwal and
                  Jy{-}yong Sohn and
                  Kangwook Lee and
                  Dimitris S. Papailiopoulos},
  title        = {Attack of the Tails: Yes, You Really Can Backdoor Federated Learning},
  booktitle    = {Neural Information Processing Systems},
  year         = {2020}
}

@inproceedings{DBLP:conf/uss/FangCJG20,
  author       = {Minghong Fang and
                  Xiaoyu Cao and
                  Jinyuan Jia and
                  Neil Zhenqiang Gong},
  title        = {Local Model Poisoning Attacks to Byzantine-Robust Federated Learning},
  booktitle    = {USENIX Security Symposium},
  pages        = {1605--1622},
  year         = {2020}
}

@inproceedings{DBLP:conf/icml/BhagojiCMC19,
  author       = {Arjun Nitin Bhagoji and
                  Supriyo Chakraborty and
                  Prateek Mittal and
                  Seraphin B. Calo},
  title        = {Analyzing Federated Learning through an Adversarial Lens},
  booktitle    = {International Conference on Machine Learning},
  volume       = {97},
  pages        = {634--643},
  year         = {2019}
}

@inproceedings{DBLP:conf/nips/ZhangJCLW23,
  author       = {Hangfan Zhang and
                  Jinyuan Jia and
                  Jinghui Chen and
                  Lu Lin and
                  Dinghao Wu},
  title        = {{A3FL:} Adversarially Adaptive Backdoor Attacks to Federated Learning},
  booktitle    = {Neural Information Processing Systems},
  year         = {2023}
}

@inproceedings{DBLP:conf/iclr/XieHCL20,
  author       = {Chulin Xie and
                  Keli Huang and
                  Pin{-}Yu Chen and
                  Bo Li},
  title        = {{DBA:} Distributed Backdoor Attacks against Federated Learning},
  booktitle    = {International Conference on Learning Representations},
  year         = {2020}
}

@inproceedings{DBLP:conf/icml/ZhangPSYMMR022,
  author       = {Zhengming Zhang and
                  Ashwinee Panda and
                  Linyue Song and
                  Yaoqing Yang and
                  Michael W. Mahoney and
                  Prateek Mittal and
                  Kannan Ramchandran and
                  Joseph Gonzalez},
  title        = {Neurotoxin: Durable Backdoors in Federated Learning},
  booktitle    = {International Conference on Machine Learning},
  volume       = {162},
  pages        = {26429--26446},
  year         = {2022}
}

@inproceedings{DBLP:conf/sp/LiYHLWFS23,
  author       = {Haoyang Li and
                  Qingqing Ye and
                  Haibo Hu and
                  Jin Li and
                  Leixia Wang and
                  Chengfang Fang and
                  Jie Shi},
  title        = {3DFed: Adaptive and Extensible Framework for Covert Backdoor Attack
                  in Federated Learning},
  booktitle    = {IEEE Security and Privacy},
  pages        = {1893--1907},
  year         = {2023}
}

@inproceedings{DBLP:conf/iclr/MadryMSTV18,
  author       = {Aleksander Madry and
                  Aleksandar Makelov and
                  Ludwig Schmidt and
                  Dimitris Tsipras and
                  Adrian Vladu},
  title        = {Towards Deep Learning Models Resistant to Adversarial Attacks},
  booktitle    = {International Conference on Learning Representations},
  year         = {2018}
}

@article{DBLP:journals/access/CuiDJZHY23,
  author       = {Chi Cui and
                  Haiping Du and
                  Zhijuan Jia and
                  Xiaofei Zhang and
                  Yuchu He and
                  Yanyan Yang},
  title        = {Data Poisoning Attacks With Hybrid Particle Swarm Optimization Algorithms
                  Against Federated Learning in Connected and Autonomous Vehicles},
  journal      = {{IEEE} Access},
  volume       = {11},
  pages        = {136361--136369},
  year         = {2023}
}

@article{DBLP:journals/access/AlmutairiB25,
  author       = {Suzan Almutairi and
                  Ahmed Barnawi},
  title        = {Exploring the Limitations of Federated Learning: {A} Novel Wasserstein
                  Metric-Based Poisoning Attack on Traffic Sign Classification},
  journal      = {{IEEE} Access},
  volume       = {13},
  pages        = {118264--118280},
  year         = {2025}
}

@article{DBLP:journals/access/MasudaKKTH23,
  author       = {Hiroki Masuda and
                  Kentaro Kita and
                  Yuki Koizumi and
                  Junji Takemasa and
                  Toru Hasegawa},
  title        = {Byzantine-Resilient Secure Federated Learning on Low-Bandwidth Networks},
  journal      = {{IEEE} Access},
  volume       = {11},
  pages        = {51754--51766},
  year         = {2023}
}

@article{DBLP:journals/access/GhaziFN25,
  author       = {Saeedeh Ghazi and
                  Saeed Farzi and
                  Amirhossein Nikoofard},
  title        = {Federated Learning for All: {A} Reinforcement Learning-Based Approach
                  for Ensuring Fairness in Client Selection},
  journal      = {{IEEE} Access},
  volume       = {13},
  pages        = {118515--118535},
  year         = {2025}
}

@article{DBLP:journals/access/YeoL25,
  author       = {Hasung Yeo and
                  Joon{-}Woo Lee},
  title        = {Norm-Based Outlier Filtering and Consensus Aggregation for Robust
                  Federated Learning},
  journal      = {{IEEE} Access},
  volume       = {13},
  pages        = {104926--104936},
  year         = {2025}
}

@article{DBLP:journals/access/Almuseelem25,
  author       = {Waleed Almuseelem},
  title        = {Secure Latency-Aware Task Offloading Using Federated Learning and
                  Zero Trust in Edge Computing for IoMT},
  journal      = {{IEEE} Access},
  volume       = {13},
  pages        = {117808--117830},
  year         = {2025}
}

@article{ACM:journals/jisa/Hu25,
  author = {Chenyu Hu and Qiming Hu and Mingyue Zhang and Zheng Yang},
  title = {FDBA: Feature-guided Defense against Byzantine and Adaptive attacks in Federated Learning},
  journal = {Journal of Information Security and Applications},
  volume = {90},
  pages = {104035},
  year = {2025}
}

@article{DBLP:journals/corr/abs-1911-07963,
  author       = {Ziteng Sun and
                  Peter Kairouz and
                  Ananda Theertha Suresh and
                  H. Brendan McMahan},
  title        = {Can You Really Backdoor Federated Learning?},
  journal      = {CoRR},
  volume       = {abs/1911.07963},
  year         = {2019}
}

@article{DBLP:journals/fttcs/DworkR14,
  author       = {Cynthia Dwork and
                  Aaron Roth},
  title        = {The Algorithmic Foundations of Differential Privacy},
  journal      = {Foundations and Trends in Theoretical Computer Science},
  volume       = {9},
  number       = {3-4},
  pages        = {211--407},
  year         = {2014}
}

@article{DBLP:journals/corr/abs-2011-01767,
  author       = {Chen Wu and
                  Xian Yang and
                  Sencun Zhu and
                  Prasenjit Mitra},
  title        = {Mitigating Backdoor Attacks in Federated Learning},
  journal      = {CoRR},
  volume       = {abs/2011.01767},
  year         = {2020}
}

@article{DBLP:journals/tsp/PillutlaKH22,
  author       = {Krishna Pillutla and
                  Sham M. Kakade and
                  Za{\"{\i}}d Harchaoui},
  title        = {Robust Aggregation for Federated Learning},
  journal      = {IEEE Transactions on Signal Processing},
  volume       = {70},
  pages        = {1142--1154},
  year         = {2022}
}

@inproceedings{DBLP:conf/icml/YinCRB18,
  author       = {Dong Yin and
                  Yudong Chen and
                  Kannan Ramchandran and
                  Peter L. Bartlett},
  title        = {Byzantine-Robust Distributed Learning: Towards Optimal Statistical
                  Rates},
  booktitle    = {International Conference on Machine Learning},
  volume       = {80},
  pages        = {5636--5645},
  year         = {2018}
}

@inproceedings{DBLP:conf/aaai/OzdayiKG21,
  author       = {Mustafa Safa {\"{O}}zdayi and
                  Murat Kantarcioglu and
                  Yulia R. Gel},
  title        = {Defending against Backdoors in Federated Learning with Robust Learning
                  Rate},
  booktitle    = {The Association for the Advancement of Artificial Intelligence},
  pages        = {9268--9276},
  year         = {2021}
}

@inproceedings{DBLP:conf/ndss/CaoF0G21,
  author       = {Xiaoyu Cao and
                  Minghong Fang and
                  Jia Liu and
                  Neil Zhenqiang Gong},
  title        = {FLTrust: Byzantine-robust Federated Learning via Trust Bootstrapping},
  booktitle    = {Network and Distributed System Security Symposium},
  year         = {2021}
}

@inproceedings{DBLP:conf/iccv/HuangLCS023,
  author       = {Siquan Huang and
                  Yijiang Li and
                  Chong Chen and
                  Leyu Shi and
                  Ying Gao},
  title        = {Multi-metrics adaptively identifies backdoors in Federated learning},
  booktitle    = {International Conference on Computer Vision},
  pages        = {4629--4639},
  year         = {2023}
}

@inproceedings{DBLP:conf/raid/FungYB20,
  author       = {Clement Fung and
                  Chris J. M. Yoon and
                  Ivan Beschastnikh},
  title        = {The Limitations of Federated Learning in Sybil Settings},
  booktitle    = {International Symposium on Research in Attacks, Intrusions and Defenses},
  pages        = {301--316},
  year         = {2020}
}

@article{DBLP:journals/tifs/MaMMLD22,
  author       = {Zhuoran Ma and
                  Jianfeng Ma and
                  Yinbin Miao and
                  Yingjiu Li and
                  Robert H. Deng},
  title        = {ShieldFL: Mitigating Model Poisoning Attacks in Privacy-Preserving
                  Federated Learning},
  journal      = {IEEE Transactions on Information Forensics and Security},
  volume       = {17},
  pages        = {1639--1654},
  year         = {2022}
}

@inproceedings{DBLP:conf/uss/NguyenRCYMFMMMZ22,
  author       = {Thien Duc Nguyen and
                  Phillip Rieger and
                  Huili Chen and
                  Hossein Yalame and
                  Helen M{\"{o}}llering and
                  Hossein Fereidooni and
                  Samuel Marchal and
                  Markus Miettinen and
                  Azalia Mirhoseini and
                  Shaza Zeitouni and
                  Farinaz Koushanfar and
                  Ahmad{-}Reza Sadeghi and
                  Thomas Schneider},
  title        = {{FLAME:} Taming Backdoors in Federated Learning},
  booktitle    = {USENIX Security Symposium},
  pages        = {1415--1432},
  year         = {2022}
}

@article{DBLP:journals/tdsc/MuCSLCZM24,
  author       = {Xutong Mu and
                  Ke Cheng and
                  Yulong Shen and
                  Xiaoxiao Li and
                  Zhao Chang and
                  Tao Zhang and
                  Xindi Ma},
  title        = {FedDMC: Efficient and Robust Federated Learning via Detecting Malicious
                  Clients},
  journal      = {IEEE Transactions on Dependable and Secure Computing},
  volume       = {21},
  number       = {6},
  pages        = {5259--5274},
  year         = {2024}
}

@article{DBLP:journals/tifs/ZhangZSGCSY24,
  author       = {Jiale Zhang and
                  Chengcheng Zhu and
                  Xiaobing Sun and
                  Chunpeng Ge and
                  Bing Chen and
                  Willy Susilo and
                  Shui Yu},
  title        = {FLPurifier: Backdoor Defense in Federated Learning via Decoupled Contrastive
                  Training},
  journal      = {IEEE Transactions on Information Forensics and Security},
  volume       = {19},
  pages        = {4752--4766},
  year         = {2024}
}

@article{DBLP:journals/corr/KonecnyMYRSB16,
  author       = {Jakub Kone{\v{c}}n{\'y} and
                  H. Brendan McMahan and
                  Felix X. Yu and
                  Peter Richt{\'{a}}rik and
                  Ananda Theertha Suresh and
                  Dave Bacon},
  title        = {Federated Learning: Strategies for Improving Communication Efficiency},
  journal      = {CoRR},
  volume       = {abs/1610.05492},
  year         = {2016}
}

@article{DBLP:journals/tist/YangLCT19,
  author       = {Qiang Yang and
                  Yang Liu and
                  Tianjian Chen and
                  Yongxin Tong},
  title        = {Federated Machine Learning: Concept and Applications},
  journal      = {ACM Transactions on Intelligent Systems and Technology},
  volume       = {10},
  number       = {2},
  pages        = {12:1--12:19},
  year         = {2019}
}

@article{DBLP:journals/cn/ZhangWLY25,
  author       = {Heng{-}Ru Zhang and
                  Ke{-}Xiong Wang and
                  Xiang{-}Yu Liang and
                  Yi{-}Fan Yu},
  title        = {{DUPS:} Data poisoning attacks with uncertain sample selection for
                  federated learning},
  journal      = {Comput Networks},
  volume       = {256},
  pages        = {110909},
  year         = {2025}
}

@article{DBLP:journals/tifs/BaiZZYH25,
  author       = {Li Bai and
                  Xinwei Zhang and
                  Sen Zhang and
                  Qingqing Ye and
                  Haibo Hu},
  title        = {ProVFL: Property Inference Attacks Against Vertical Federated Learning},
  journal      = {IEEE Transactions on Information Forensics and Security},
  volume       = {20},
  pages        = {6529--6543},
  year         = {2025}
}

@inproceedings{DBLP:conf/ccnc/SanonRLS24,
  author       = {Sogo Pierre Sanon and
                  Rekha Reddy and
                  Christoph Lipps and
                  Hans Dieter Schotten},
  title        = {DDoS Attacks in Communication: Analysis and Mitigation of Unreliable
                  Clients in Federated Learning},
  booktitle    = {IEEE Consumer Communications and Networking Conference},
  pages        = {986--989},
  year         = {2024}
}

@inproceedings{DBLP:conf/aistats/FraboniVL21,
  author       = {Yann Fraboni and
                  Richard Vidal and
                  Marco Lorenzi},
  title        = {Free-rider Attacks on Model Aggregation in Federated Learning},
  booktitle    = {International Conference on Artificial Intelligence and Statistics},
  volume       = {130},
  pages        = {1846--1854},
  year         = {2021}
}

@article{DBLP:journals/tifs/MaHWQWM25,
  author       = {Zhuoran Ma and
                  Xinyi Huang and
                  Zhuzhu Wang and
                  Zhan Qin and
                  Xiangyu Wang and
                  Jianfeng Ma},
  title        = {FedGhost: Data-Free Model Poisoning Enhancement in Federated Learning},
  journal      = {IEEE Transactions on Information Forensics and Security},
  volume       = {20},
  pages        = {2096--2108},
  year         = {2025}
}

@article{DBLP:journals/tifs/YangMLLLKLD25,
  author       = {Li Yang and
                  Yinbin Miao and
                  Ziteng Liu and
                  Zhiquan Liu and
                  Xinghua Li and
                  Da Kuang and
                  Hongwei Li and
                  Robert H. Deng},
  title        = {Enhanced Model Poisoning Attack and Multi-Strategy Defense in Federated
                  Learning},
  journal      = {IEEE Transactions on Information Forensics and Security},
  volume       = {20},
  pages        = {3877--3892},
  year         = {2025}
}

@article{DBLP:journals/tbd/XiaoTLJL24,
  author       = {Xiong Xiao and
                  Zhuo Tang and
                  Chuanying Li and
                  Bingting Jiang and
                  Kenli Li},
  title        = {{SBPA:} Sybil-Based Backdoor Poisoning Attacks for Distributed Big
                  Data in AIoT-Based Federated Learning System},
  journal      = {IEEE Transactions on Big Data},
  volume       = {10},
  number       = {6},
  pages        = {827--838},
  year         = {2024}
}

@article{DBLP:journals/access/ChenFW25,
  author       = {Xiao Chen and
                  Chao Feng and
                  Shaohua Wang},
  title        = {{AIDFL:} An Information-Driven Anomaly Detector for Data Poisoning
                  in Decentralized Federated Learning},
  journal      = {{IEEE} Access},
  volume       = {13},
  pages        = {50017--50031},
  year         = {2025}
}

@inproceedings{DBLP:conf/cvpr/XieFG25,
  author       = {Yueqi Xie and
                  Minghong Fang and
                  Neil Zhenqiang Gong},
  title        = {Model Poisoning Attacks to Federated Learning via Multi-Round Consistency},
  booktitle    = {IEEE Conference on Computer Vision and Pattern Recognition},
  pages        = {15454--15463},
  year         = {2025}
}

@article{DBLP:journals/iotj/WangCZHWDLL25,
  author       = {Xinxin Wang and
                  Jing Chen and
                  Zijun Zhang and
                  Kun He and
                  Zongru Wu and
                  Ruiying Du and
                  Qiao Li and
                  Gongshen Liu},
  title        = {Transferable and Robust Dynamic Adversarial Attack Against Object
                  Detection Models},
  journal      = {IEEE Internet of Things Journal},
  volume       = {12},
  number       = {11},
  pages        = {16171--16180},
  year         = {2025}
}

@inproceedings{DBLP:conf/uss/PanZWXJY20,
  author       = {Xudong Pan and
                  Mi Zhang and
                  Duocai Wu and
                  Qifan Xiao and
                  Shouling Ji and
                  Min Yang},
  title        = {Justinian's GAAvernor: Robust Distributed Learning with Gradient Aggregation
                  Agent},
  booktitle    = {USENIX Security Symposium},
  pages        = {1641--1658},
  year         = {2020}
}

@article{DBLP:journals/access/AliHJ25,
  author       = {Asad Ali and
                  Jianjun Huang and
                  Ayesha Jabbar},
  title        = {Recent Advances in Federated Learning for Connected Autonomous Vehicles:
                  Addressing Privacy, Performance, and Scalability Challenges},
  journal      = {{IEEE} Access},
  volume       = {13},
  pages        = {80637--80665},
  year         = {2025}
}

@article{DBLP:journals/access/NairNRN25,
  author       = {Divya G. Nair and
                  C. V. Aswartha Narayana and
                  K. Jaideep Reddy and
                  Jyothisha J. Nair},
  title        = {FedHSP: {A} Robust Federated Learning Framework Coherently Addressing
                  Heterogeneity, Security, and Performance Challenges},
  journal      = {{IEEE} Access},
  volume       = {13},
  pages        = {77049--77063},
  year         = {2025}
}

\end{document}